\documentclass[11pt]{article}

\usepackage[preprint]{acl}

\usepackage{times}
\usepackage{latexsym}
\usepackage{amsmath} 
\usepackage{amssymb}
\usepackage{booktabs}
\usepackage{float}
\usepackage{authblk}  
\usepackage{longtable}
\usepackage{stfloats}
\usepackage{makecell} 
\usepackage{graphicx}    
\usepackage{caption}     
\usepackage{multirow}
\usepackage[T1]{fontenc}

\usepackage[utf8]{inputenc}

\usepackage{microtype}

\usepackage{inconsolata}

\usepackage{graphicx}

\title{MedReflect: Teaching Medical LLMs to Self-Improve via Reflective Correction}
\makeatletter
\renewcommand{\@fnsymbol}[1]{\ensuremath{\ifcase#1\or *\or \dagger\or \ddagger\or
   \mathsection\or \mathparagraph\or \|\or **\or \dagger\dagger\else\@ctrerr\fi}}
\makeatother

\author{
    Yue Huang\textsuperscript{1}\thanks{These authors contributed equally to this work}, 
    Yanyuan Chen\textsuperscript{2}\textsuperscript{*}, 
    Dexuan Xu\textsuperscript{1}, 
    Chenzhuo Zhao\textsuperscript{1}, 
    Weihua Yue\textsuperscript{3}, 
    Yu Huang\textsuperscript{1}\\
    \textsuperscript{1}Peking University \\
    \textsuperscript{2}University of Virginia \\
    \textsuperscript{3}Peking University Sixth Hospital \\
}
\begin{document}
\maketitle
\begin{abstract}
Medical problem-solving demands expert knowledge and intricate reasoning. Recent studies of large language models (LLMs) attempt to ease this complexity by introducing external knowledge verification through retrieval-augmented generation or by training on reasoning datasets. However, these approaches suffer from drawbacks such as retrieval overhead and high annotation costs, and they heavily rely on substituted external assistants to reach limited performance in medical field. In this paper, we introduce MedReflect, a generalizable framework designed to inspire LLMs with a physician‑like reflective thinking mode. MedReflect generates a single‑pass reflection chain that includes initial hypothesis generation, self‑questioning, self‑answering and decision refinement. This self-verified and self-reflective nature releases large language model's latent capability in medical problem-solving without external retrieval or heavy annotation. We demonstrate that MedReflect enables cost-efficient medical dataset construction. With only a minimal subset of randomly sampled training examples and lightweight fine-tuning, this approach achieves notable absolute accuracy improvements across a series of medical benchmarks while significantly cutting annotation requirements. Our results provide evidence that LLMs can learn to solve specialized medical problems via self-reflection and self-improvement, reducing reliance on external supervision and extensive task-specific fine-tuning data.

\end{abstract}

\section{Introduction}
\begin{figure}
    \centering
    \includegraphics[width=1\linewidth]{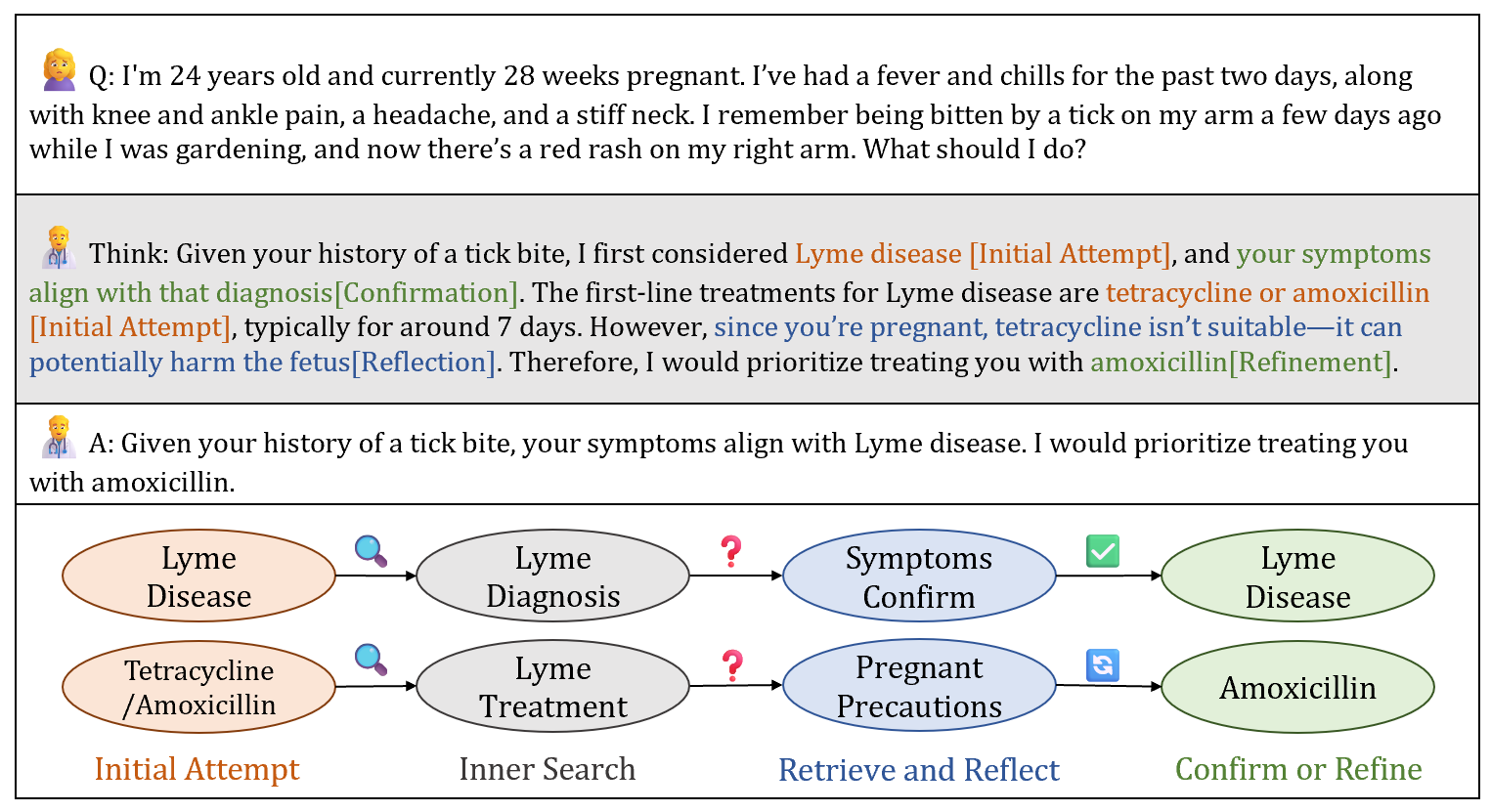}
    \caption{Example of the Physician Clinical Reasoning and Self-Correction Process Simulated by MedReflect. This figure demonstrates, through a specific clinical case, how the MedReflect framework simulates the human physician's cognitive loop of "Hypothesis-Verification-Reflection-Refinement.}
    \label{fig:Example}
\end{figure}

Recent progress of large language models (LLMs) has showcased their immense potential in medical tasks \cite{LIN2025100868,xu2025knowledge}. Despite this progress, deploying LLMs in professional medical scenarios offers unique challenges compared to general domains. Medical decision-making is generally narrower in scope and involves complex scenarios, which necessitate meticulous thinking to ensure more reliable answers. In addition, specialized medical terminology and intricate clinical narratives may increase hallucinated or unreliable outputs~\cite{dou-etal-2024-detection}. 

Existing efforts primarily improve medical LLMs by injecting external medical knowledge. Retrieval-Augmented Generation method (RAG), for instance, consults external sources to verify and refine the generated answers~\cite{wu2024medicalgraphragsafe,lu2025doctorragmedicalragfusing}. However, this method heavily relies on additional external knowledge and incurs extra storage and retrieval overhead. 

Recently, reasoning techniques have been explored to guide LLMs through predefined, structured sequence of analytical steps during generation \cite{kwon2024large}. For instance, MedReason\cite{wu2025medreason} leverages structured knowledge graphs to synthesize reasoning chains and construct Chain-of-Thought (CoT) data. However, these methods rely on carefully curated datasets to build reasoning paths and require domain experts to predefine comprehensive medical reasoning processes, substantially increasing annotation costs. To alleviate this burden, HuatuoGPT-o1 \cite{chen2024huatuogpt} use an auxiliary LLM verifier to assess model-generated answers and guide the correction of errors through external feedback signals, highlighting the potential of language models to self-correct under the guidance of external signals. 

However, these approaches substitute external mechanisms for the model’s internal capabilities of knowledge localization and reasoning planning. This leads to a fundamental question: Can medical language models learn to generate hypotheses, retrieve relevant medical knowledge, and perform self-verification and correction—all within a single generation process?  The motivation for raising this question is that LLMs already possess extensive medical knowledge during pretraining \cite{vladika2024medreqal}. As discussed above, when faced with complex medical problems, LLMs can be guided by external verifiers to rediscover the correct reasoning path. This resembles clinical reasoning in practice: careful deliberation that involves formulating an initial hypothesis, retrieving relevant medical knowledge, and then validating or revising conclusions through iterative reasoning, which reflects the meticulous thinking required in medical decision-making.  An illustrative example is shown in Figure \ref{fig:Example}.


In this work, we systematize the key components of the medical decision-making process and propose MedReflect. Our key idea is to equip the model with a directional reflection-and-correction process. We design a structured reflection mechanism that guides the model to generate and answer its own questions during reasoning, which injects direction into reflection. Through this self-questioning–self-answering procedure, the model explicitly probes uncertain assumptions, retrieves relevant medical knowledge on its own, and revises intermediate or final decisions accordingly. As a result, MedReflect does not rely on external knowledge retrieval methods, and instead, it uses self-generated questions and answers to perform internal retrieval and correction within a single generation. Experiments demonstrate that our method yields improvement on medical benchmarks. Our contributions are as follows:

\begin{itemize}
    \item We propose \textbf{MedReflect}, a reflection-and-correction mechanism to enhance the depth and quality of reasoning in medical LLMs.
    
    \item We develop a practical approach that leverages LLMs to construct a low-cost medical reflection  dataset.
    
    \item Experiments on multiple medical QA benchmarks show that MedReflect consistently improves accuracy. The results demonstrate that reflective supervision effectively teaches models to self-reflect and self-correct during generation, outperforming existing chain-of-thought training methods in both performance and training efficiency.
\end{itemize}

\section{Related Works}

\paragraph{Medical LLMs.} 
Prior work has extensively explored developing medical-specific LLMs to excel in the medical domain~\cite{chen2024huatuogpt,chen2023huatuogpt,team2025gemma,labrak2024biomistral,zhang2024ultramedical}. While promising, applying LLMs to complex medical cases remains challenging~\cite{rios2024evaluation}, with persistent concerns regarding ethics and hallucinations~\cite{soffer2025pitfalls}. To address diagnostic capabilities, \citet{liu2025medical} proposes a generalist medical LLM for diagnostic reasoning across specialties. Several approaches leverage external data or specialized training to bridge these gaps: UltraMedical~\cite{zhang2024ultramedical} comprises 410K instruction-following examples, while BioMistral~\cite{labrak2024biomistral} utilizes PubMed Central for continued pre-training. \citet{jeong2024improving} further integrates retrieval with self-reflection to enhance reliability in biomedical tasks. 
Recent research has increasingly focused on structured and factual reasoning. MedReason~\cite{wu2025medreason} leverages knowledge graphs to elicit factual reasoning steps, and MedFact-R1~\cite{li2025medfact} employs pseudo-label augmentation to bolster factual medical reasoning. Similarly, \citet{chen2025evaluating} investigates improving syndrome differentiation thinking. To enable self-improvement, MuSeR~\cite{zhou2025enhancing} targets medical context-awareness via multifaceted self-refinement, while HuatuoGPT-o1~\cite{chen2024huatuogpt} advances medical reasoning through a two-stage approach combining SFT, verifiable medical problems, and reinforcement learning.

\paragraph{Developing Models for Self-reflection and Self-correction.} 
Self-reflection has emerged as a critical mechanism for mitigating hallucinations~\cite{ji2023towards} and improving reasoning, although its effectiveness is contingent on specific conditions regarding model capability and task difficulty~\cite{kamoi2024can}. Several works have studied backtracking and search as forms of self-correction~\cite{ye2024physics,qin2025backtrack}. Notably, the Stream of Search (SoS) framework~\cite{gandhi2024stream} enables models to self-correct by searching within language without external components, a phenomenon also explored in recent work on the emergence of thinking processes in LLMs~\cite{ye2025emergence}. 
Beyond search, critique-based methods have gained traction. \citet{xi2024enhancing} trains an expert critique model using step-level feedback to supervise the reasoning process. Building on this, ReflectEvo~\cite{li2025reflectevo} explores iterative self-reflection to improve meta-introspection, while SaySelf~\cite{xu2024sayself} teaches LLMs to express confidence calibration within their self-reflective rationales. Additionally, \citet{qu2024recursive} introduces iterative fine-tuning to alter responses after unsuccessful attempts. More recently, \citet{zhu2025emergence} has begun to probe the underlying mechanisms for controlling and modulating these self-reflection behaviors.

\section{Problem Setup and Preliminaries}
\begin{figure*}[t]
    \centering
    \includegraphics[width=1\linewidth, trim=0 0 0 0, clip]{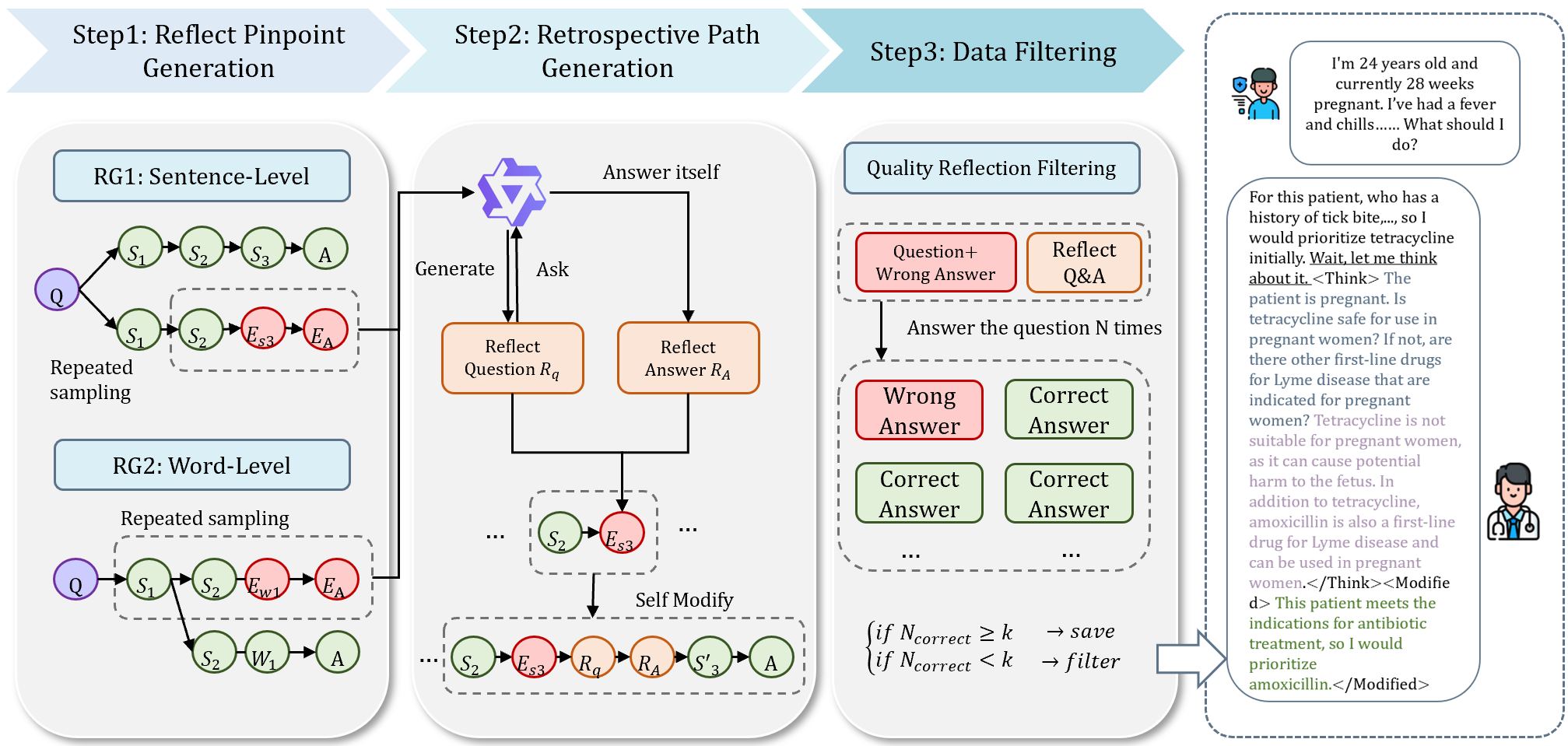}
    \caption{Data Construction Framework}
    \label{fig:data-construct}
\end{figure*}

We aim to explore whether medical LLMs can generate hypotheses, retrieve, perform self-verify and correction within a single generation process. To achieve this, we introduce MedReflect, focused on teaching LLMs to learn the reflect mode instead of directly injecting knowledge or improving their reasoning skills. The overview of the data generation of MedReflect is shown in Figure~\ref{fig:data-construct}.

Given a medical question $Q$ and its corresponding response trajectory $ T = [S_1, S_2, \ldots, S_n,A] $, the LLM is prompted to regenerate either a selected step $ S_i $ or the entire answer. This regeneration may introduce errors, resulting in an erroneous trajectory  or answer to the whole question $ T_{er} = [S_1, S_2, \ldots, E_i, \ldots,E_A] $. Guided by the error $ E_i $ and the erroneous trajectory $ T_{er} $, LLM then generates a targeted reflection question $ R_{q_i} $ along with its answer $ R_{a_i} $. Finally, the original question $ Q $, the erroneous trajectory $ T_{er} $, and the reflection pair $(R_{q_i}, R_{a_i})$ are provided to LLM to facilitate error correction. Upon successful correction, the original correct step $ S_i $ and the reflection pair are incorporated into the trajectory, yielding the final reflective trajectory $ T_{reflect} = [S_1, S_2, \ldots, E_i, R_{q_i}, R_{a_i}, S_i, \ldots,A] $.

\section{Methodology}

\subsection{Datasource and Construction Model}
We utilize two publicly available medical datasets: ChatDoctor \cite{li2023chatdoctormedicalchatmodel} and training split of MedMCQA \cite{pal2022medmcqa}. ChatDoctor contains 100k real-world conversations between patients and doctors sourced from HealthCareMagic.com. MedMCQA is a comprehensive multiple-choice question-answering dataset created specifically from real medical entrance exam questions. We used Qwen2.5-32B-Instruct to complete the entire data construction process.

\subsection{Reflect Pinpoint Generation}
\begin{figure*}
    \centering
    \includegraphics[width=1\linewidth]{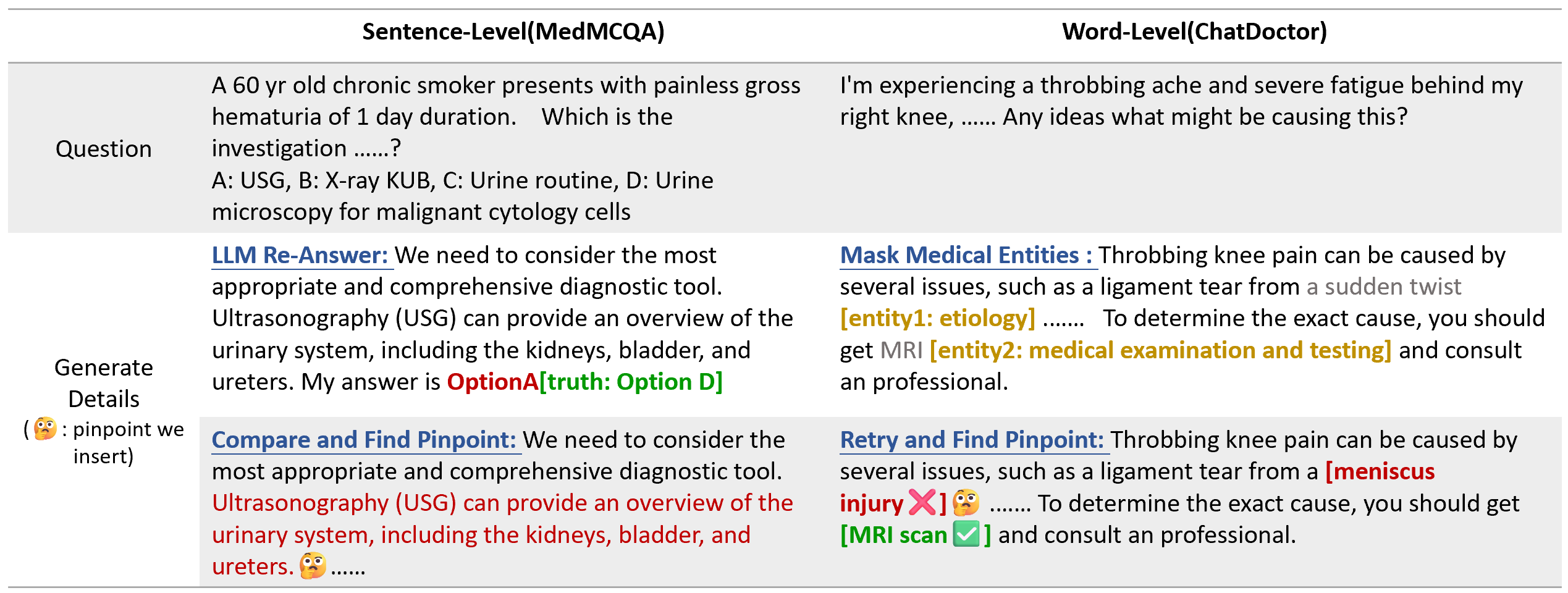}
    \caption{Examples of Pinpoint Generation}
    \label{fig:pinpoint}
\end{figure*}

To better control the diversity of generated pinpoints, we designed multiple reflective pinpoint generation pathways(RG), including reflective processes functioning at two different levels of detail. Figure \ref{fig:pinpoint} shows examples of two different levels.

\paragraph{RG1: Sentence Level.} This approach primarily focuses on the overall rewriting of reasoning sentences generated by the LLM. We utilize this method to construct training data derived from the multi-choice MedMCQA dataset. Specifically, we perform repeated sampling by prompting the LLM to respond to the original question $Q$ multiple times, generating new answers that include their reasoning processes. For each sampled response, we employ heuristic pattern matching to extract the multi-choice decision statement and compare it with the ground truth. If the extracted decision answer $A'$ is incorrect, we locate the specific sentence within the response that led to the wrong option, designating it as a reflection pinpoint $E_{si}$.
\paragraph{RG2: Word Level}
This approach focuses on medical texts rich in entities. We used this method to construct the data of the consultation dataset ChatDoctor. We extract entity $W_i$ from medical consultation reasoning sentence $S_i$ and mask it with its corresponding entity type, such as diseases, etiology, and treatment. The LLM is prompted to predict and fill in the masked entity multiple times. If an attempt by the LLM is assessed as incorrect and such errors occur frequently, the sentence containing this entity serves as our reflection pinpoint $E_{wi}$. In contrast, responses that are assessed as correct will not be treated as such pinpoints. At this level of detail, a single physician’s response can produce one to three pinpoints for reflection.

\subsubsection{Retrospective Path Generation}

\paragraph{Reflect QA Generation}
To create reflection questions $R_q$ and their matching answers $R_a$, we provide LLM with its original answer and emphasize that its answer was wrong. Through targeted prompting, we guide LLM to formulate a focused reflective question $R_q$. The goal of this question is to help LLM identify the key knowledge components needed to develop a better solution.

Afterward, the LLM is then instructed to answer the reflection question using only its own knowledge, without referring to the original question or adding extra details. The purpose of this method is to engage and utilize LLM's internal knowledge, promoting thorough self-awareness and reflection.
\paragraph{Modification Based on Reflection}
Using $Q$, $R_q$ and $R_a$  as input, we guide the LLM to generate the modified statement $M$.

(1) For RG1, LLM regenerates sentences $S'_i$ to replace original sentences with expression flaws, performing adaptive optimization to resolve any contextual coherence issues caused by the new sentences.

(2) For RG2, LLM tries to use the new response word $W'_i$ to substitute inaccurate vocabulary.

 These modified parts will be inspected in the next step to ensure they are qualified.
\subsubsection{Data Filtering}
To guarantee data quality, we first filtered out samples with insufficient reasoning or irrelevant content.  We then implemented a secondary validation loop to mitigate confirmation bias by feeding the generated reflections ($R_q, R_a$) and the initial erroneous trajectory back into the model.  A reflection was deemed valid only if it guided the model to the objective ground truth—specifically, selecting the correct option for RG1 or restoring the masked entity for RG2.  Finally, to exclude stochastic reasoning, we enforced a robustness constraint based on repeated trials. We conducted $[k]$ independent inference trials for each validated sample and calculated the success rate. Only instances exceeding a consistency threshold of $\tau = 0.8$ were retained This yielded a final dataset of 36,413 medical consultation records and 21,107 multiple-choice questions.

\subsection{Training Strategy}

Following the construction of the generated reflection dataset, we investigate strategies to leverage these samples to enhance the model's reasoning capabilities. We formalize the medical reasoning process as a sequential decision-making problem within a discrete semantic space. In conventional foundation models, the policy is often dominated by next-token prediction objectives derived from pre-training corpus statistics. This approach frequently results in logical disconnects or hallucinations when addressing complex medical cases. As illustrated in previous works, merely extending the Chain of Thought (CoT) is analogous to performing a directionless random walk on a semantic manifold; it induces divergent thinking but relies on stochastic retries to identify a superior solution.

To address this limitation, the \textbf{MedReflect} framework introduces a reflection operator designed to provide directed self-correction to the otherwise undirected reasoning process. This is formally defined as:
\begin{equation}
    \tau_{reflect} = \mathcal{R}(\tau_{err}, \mathcal{K}_{int}) \Rightarrow (R_q, R_a)
    \label{eq:reflection_operator}
\end{equation}
Here, $\tau_{err}$ represents the initial trajectory containing error breakpoints, and $\mathcal{K}_{int}$ denotes the model's implicit medical knowledge. The reflection pair $(R_q, R_a)$ serves as a critical control signal to rectify $\tau_{err}$. Specifically, $R_q$ facilitates \textit{error attribution} by compelling the model to pinpoint knowledge deficits, while $R_a$ leverages $\mathcal{K}_{int}$ to generate a rectified factual basis.

To instantiate this theoretical operator within the language model, we structure the reasoning process as a token-augmented trajectory. We introduce four special tokens---\texttt{<Think>}, \texttt{</Think>}, \texttt{<Modified>}, and \texttt{</Modified>}---to explicitly delineate the semantic boundaries of the reflection process. Consequently, the training instance is transformed from a standard input-output pair into a \textit{reflective sequence} $\mathbf{y}_{seq}$, constructed as:
\begin{equation}
\begin{split}
    \mathbf{y}_{seq} = [ & \tau_{err}, \texttt{<Think>}, R_q, R_a, \texttt{</Think>}, \\
    & \texttt{<Modified>}, \tau_{corrected}, \texttt{</Modified>}]
\end{split}
\label{eq:sequence_construction}
\end{equation}
This structure forces the model to treat reflection not as an external module, but as an intrinsic part of its generative policy.

Finally, we align the reflective medical LLM with this structured reasoning pattern through supervised fine-tuning (SFT). We interpret SFT here as behavioral cloning of the reflection operator. Given the dataset $\mathcal{D}_{\text{med}} = \{ (\mathbf{x}, \mathbf{y}_{seq}) \}$, the optimization objective is to maximize the likelihood of the augmented reflective sequence. The loss function for the model $\pi_{\theta}$ is defined as:
\begin{equation}
    \mathcal{L}_{\text{SFT}}(\theta) = -\mathbb{E}_{(\mathbf{x}, \mathbf{y}_{seq}) \sim \mathcal{D}_{\text{med}}} \left[ \sum_{t=1}^{T} \log \pi_{\theta}(y_t \mid \mathbf{x}, y_{<t}) \right]
    \label{eq:sft_loss}
\end{equation}
By minimizing this loss, the model learns to sequentially generate the error analysis and correction steps before outputting the final answer, effectively internalizing the $\tau_{reflect}$ operator into its parameters.

\section{Experiments}

\subsection{Benchmark}
We evaluate on standard medical benchmarks: MedQA(USMLE test set) \cite{jin2021disease}, MedMCQA(validation set), and PubMedQA(test set) \cite{jin2019pubmedqa}. To thoroughly assess the model's abilities, we also included medical sections from multi-domain evaluation frameworks, specifically the Health and Biology topics in MMLU-Pro \cite{wang2024mmlu}, as well as the Genetics and Molecular Biology areas in GPQA \cite{rein2024gpqa}. These benchmarks cover tasks in various aspects such as diagnosis, treatment, knowledge, and medical reasoning.
\subsection{Baselines and Models}
We fine-tuned Qwen2.5-7B-Instruct and Llama-3.1-8B-Instruct using 2k training examples each, and Qwen2.5-32B-Instruct using 30k examples; all training sets were randomly sampled from the MedReflect dataset. To strictly evaluate the efficacy of our approach, we benchmarked MedReflect against a comprehensive suite of baselines, ranging from standard open-source medical and general LLMs to closed-source models. We also explicitly compared our model against reasoning-oriented baselines (e.g., HuatuoGPT-o1) under comparable settings.
\subsection{Implementation Details}
We fine-tuned for 3 epochs with $lr = 1e-4$. Additionally, we applied Low-Rank Adaptation (LoRA) \cite{hu2022lora}  with $\alpha = 8$ to fine-tune the LLM. The experiments were carried out on 4 NVIDIA A800 GPUs. More implementation details of training can be found in appendix \ref{app:appendix_implementation}.

\subsection{Main Results}
\begin{table*}[h]
\centering
\small
\begin{tabular}{lcccccccc}
\toprule
\multirow{2}{*}{\textbf{Model}} &
\multirow{2}{*}{\textbf{MedQA}} &
\multirow{2}{*}{\textbf{MedMCQA}} &
\multirow{2}{*}{\textbf{PubMedQA}} &
\multicolumn{2}{c}{\textbf{MMLU-Pro}} &
\multicolumn{2}{c}{\textbf{GPQA}}  \\
\cmidrule(lr){5-8}
& & & & 
\textbf{Health} & 
\textbf{Biology} & 
\textbf{Genetics} & 
\textbf{\makecell{Molecular\\Biology}} \\  
\midrule
\multicolumn{8}{c}{\textit{$<=$8B Language Models}} \\  
BioMistral-7B & 45.0 & 40.2 & 66.9 & 27.4 & 49.2 & 28.6 & 38.5  \\
UltraMedical-8B & 71.1 & 58.3 & 77.4 & 55.1 & 66.7 & 41.2 & 48.4  \\
Qwen2.5-7B-Instruct & 57.0 & 55.6 & 55.6 & 50.6 & 70.2 & 36.2 & 49.7  \\
LLaMA-3.1-8B-Instruct & 58.7 & 56.0 & 75.2 & 52.7 & 64.6 & 33.8 & 46.8  \\
HuatuoGPT-o1-8B & 72.6 & 60.4 & \textbf{79.2} & 58.7 & 68.2 & 48.8 & 59.7  \\
MedReflect-8B(LLaMa) & 64.9 & 72.4 & 75.3 & 53.1 & 64.2 & 56.7 & 60.1  \\
MedReflect-7B(Qwen) & \textbf{75.5} & \textbf{77.1} & 75.3 & \textbf{62.8} & \textbf{75.8} & \textbf{65.0} & \textbf{60.3}  \\
\midrule
\multicolumn{8}{c}{\textit{$>$ 8B Language Models}} \\  
Deepseek-R1 & 90.1 & 78.8 & 77.2 & 79.2 & 90.8 & 65.0 & 75.3  \\ 
Gemini2.5-Flash & 92.0 & 79.7 & 76.2 & -- & 98.6 & -- & -- \\  
Gemini2.5-Pro & 92.6 & 71.1 & 75.8 & -- & 98.6 & -- & --\\ 
GPT-4o & 86.5 & 76.1 & 78.4 & -- & 98.4 & -- & --  \\  
GPT-o3 & \underline{93.3} & \underline{83.3} & 80.0 & -- & \underline{98.6} & -- & --  \\  
UltraMedical-70B & 82.2 & 71.8 & 78.4 & 64.8 & 71.1 & 33.8 & 62.9  \\  
OpenBioLLM-70B & 78.3 & 74.0 & 79.0 & -- & \textbf{93.8} & -- & -- \\  

Qwen2.5-72B-Instruct & 72.7 & 66.2 & 71.7 & 65.3 & 78.8 & 41.2 & 56.8 \\  
QwQ-32B-Preview & 72.3 & 65.6 & 73.7 & 62.0 & 78.1 & 37.5 & 64.5  \\  
HuatuoGPT-o1-70B & 83.3 & 73.6 & 80.6 & 71.0 & 82.8 & 56.2 & 66.5 \\  
MedReflect-32B(Qwen) & \textbf{83.5} & \textbf{80.1} & \textbf{82.3} & \textbf{78} & 90.8 & \textbf{68.3} & \textbf{70.6} \\  
\bottomrule
\end{tabular}
\caption{Model performance on biomedical QA benchmarks.}
\label{tab:compare}
\end{table*}

We conducted a comprehensive evaluation of open-source LLMs across diverse medical benchmarks, with detailed results presented in Table \ref{tab:compare}. The experimental findings indicate that MedReflect outperforms comparable open-source models on the majority of evaluated benchmarks, establishing new state-of-the-art results for models of equivalent scale, particularly in reasoning-intensive tasks.

\paragraph{MedReflect Demonstrates Robust Reasoning Capabilities}
MedReflect-7B (Qwen) achieves significant improvements over its base model and outperforms other specialized models on key benchmarks such as MedQA and MedMCQA. Notably, it exhibits superior performance on complex reasoning benchmarks. On MMLU-Pro (Health/Biology) and GPQA, MedReflect-7B surpasses all other models in its size category. MedReflect demonstrates a more balanced and robust performance across both knowledge retrieval and complex clinical reasoning tasks.

\paragraph{Scalability and Competitiveness with Proprietary Models}
Scaling up to 32B parameters, MedReflect shows outstanding performance, surpassing significantly larger open-source models. Crucially, MedReflect-32B remains highly competitive with leading proprietary models on specific tasks. While a performance gap remains on MedQA compared to the strongest proprietary systems, MedReflect-32B significantly bridges the gap between open-source and commercial SOTA models \cite{achiam2023gpt, team2024gemini, guo2025deepseek}.

\paragraph{Generalizability Across Model Architectures}
To verify the universality of our approach, we applied the reflection mechanism to the Llama-3.1-8B-Instruct architecture. The results show that MedReflect-8B (Llama) achieves a substantial gain over its base model, confirming that the reflection mechanism is model-agnostic. However, consistent with the base models' capabilities, the Qwen-based implementation (MedReflect-7B) retains an overall performance edge over the Llama-based variant.

\paragraph{The Value of Reflection}
These advancements underscore the efficacy of the reflection chain. MedReflect achieves these results using efficient supervised fine-tuning onine-tuned a curated dataset, without relying on the massive computational resources typically required for pre-training or reinforcement learning at the scale of 70B+ models. This suggests that the self-reflection mechanism efficiently unlocks the latent reasoning capabilities of LLMs in the medical domain.

\subsection{Ablation Study}

We conducted further ablation studies on MedReflect-7B across the MedQA, PubMedQA, MMLU, and GPQA datasets to investigate how the reflection mechanism and training methodologies specifically affect performance. All variants are built upon the Qwen2.5-7B-Instruct backbone and were fine-tuned on the same fixed set of 2,000 samples. These samples were randomly selected from the MedReflect dataset and subsequently reformatted for training. This controlled setup ensures that any observed performance differences stem solely from the architectural modifications to the reasoning chain. As shown in Table~\ref{tab:Ablation}, the results support the following analyses:

\begin{table*}[h]
\centering
\small
\renewcommand{\arraystretch}{1.3}
\begin{tabular}{l c c c c c c}
\toprule
\multirow{2}{*}{\textbf{Model}} &
\multirow{2}{*}{\textbf{MedQA}} &
\multirow{2}{*}{\textbf{PubMedQA}} &
\multicolumn{2}{c}{\textbf{MMLU-Pro}} &
\multicolumn{2}{c}{\textbf{GPQA}}  \\
\cmidrule(lr){4-5} \cmidrule(lr){6-7}
& & & \textbf{Health} & \textbf{Biology} & \textbf{Genetics} & \textbf{\makecell{Molecular\\Biology}} \\  
\midrule
Qwen2.5-7B-Instruct & 57.0 & 55.6 & 50.6 & 70.2 & 36.2 & 49.7 \\

\textbf{SFT} w/o Reflect & 57.3 & 63.3 & 54.3 & 72.8 & 48.3 &  55.5\\
\textbf{SFT} w/o Reflect Question & 61.8 & 67.3 & 63.2 & 73.1 & 55.0 & 58.4 \\
\textbf{SFT} w/o Reflect Answer & 63.7 & 68.3 & 57.9 & 73.3 & 55.0 & 57.3 \\
MedReflect-7B & \textbf{75.5} & \textbf{75.3} & \textbf{62.8} & \textbf{75.8} & \textbf{65.0} & \textbf{60.3} \\

\bottomrule
\end{tabular}
\caption{Performance comparison across biomedical QA datasets using various training strategies.}
\label{tab:Ablation}
\end{table*}

\paragraph{Directional reflection significantly outperforms blind correction}
We compared MedReflect with a fine-tuning strategy that excludes the reflection step (SFT w/o Reflect). It removes the intermediate reflection process, essentially forcing the model to learn to jump directly from an error to the correct answer.

Experimental results demonstrate that MedReflect exhibits significant advantages across all benchmarks. The \textbf{SFT w/o Reflect} baseline resembles a form of random retry within the semantic space; while the model observes the correct outcome, it lacks the logical pathway to deduce the correct state from the erroneous one. While \textbf{SFT w/o Reflect} enhances performance through domain knowledge injection and direct answer supervision, MedReflect achieves superior results by equipping the model with an autonomous error-correction mechanism during the reasoning process. Furthermore, MedReflect does not simply memorize correction data but rather employs a diagnostic thinking pattern analogous to that of human physicians.

\paragraph{The integrity of the reflection chain is crucial}
To deconstruct the functional necessity of different components within the reflection chain, we compared three training configurations, all deriving from the same 2,000-sample dataset:

(1) \textbf{SFT w/o Reflect Answer}: Retains only the reflection questions ($R_q$), removing the detailed reflective answers.

(2) \textbf{SFT w/o Reflect Question}: Retains only the reflection answers ($R_a$), removing the guiding self-inquiry.

(3) \textbf{MedReflect}: Includes the full pair of reflection questions and answers ($R_q, R_a$).

The experimental data indicates that the absence of any single component leads to performance degradation, although configurations retaining partial reflection chains still outperform the baseline without reflection (\textbf{SFT w/o Reflect}). The reflection question ($R_q$) serves the role of error attribution, explicitly guiding attention toward potential defects in the reasoning chain and setting a clear direction for correction. The reflection answer ($R_a$) serves the role of knowledge retrieval and consolidation, utilizing a declarative tone to invoke the internal knowledge base and provide a factual basis for the correction. Therefore, the complete MedReflect mechanism constructs a closed cognitive loop, where the combination of $R_q$ and $R_a$ plays a critical role in effectively enhancing the model's reasoning capabilities.

\subsection{Analysis}

\subsubsection{Data Efficiency and Marginal Utility}
\begin{figure}
    \centering
    \includegraphics[width=1\linewidth]{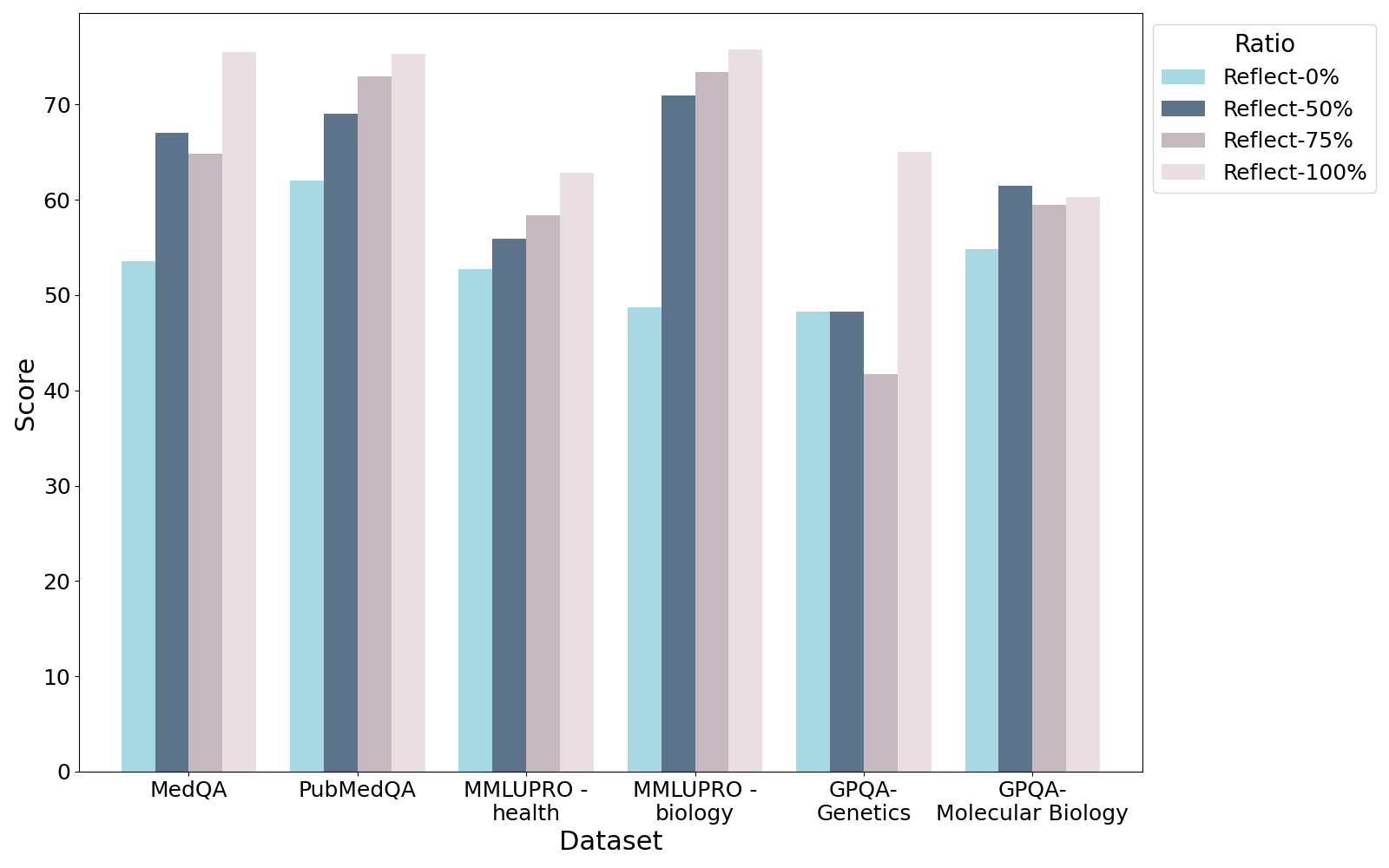}
    \caption{Results of the Reflection Data Proportion Experiment}
    \label{fig:ratio}
\end{figure}

To investigate the marginal utility of reflection data during training, we conducted experiments with four different proportions of reflection data (0\%, 50\%, 75\%, and 100\%) while maintaining a constant total volume of 2,000 training samples. As shown in Figure \ref{fig:ratio}, the results reveal a significant positive correlation between the proportion of reflection data and model performance.

\paragraph{Significant Marginal Returns of Reflection Data}
As the proportion of reflection data in the training set increases, the model's accuracy across the vast majority of medical and biological reasoning tasks demonstrates a steady upward trend. Although minor fluctuations are observed in isolated datasets, the overall trajectory indicates that a high proportion of reflection data effectively activates the model's deep reasoning capabilities. Notably, this benefit is maximized when the reflection proportion reaches 100\%.

\paragraph{Efficiency Analysis}
Under the constraint of a constant total sample size, we progressively replaced standard Correct Data with Reflection Data containing detailed reasoning processes. The results show that this substitution yields significant performance gains. This implies that, per unit data sample, data containing reflection processes possesses higher information density and training efficiency. By explicitly demonstrating error correction and logical deduction, reflection data compels the model to learn reflective thinking rather than mere rote memorization. 

\subsubsection{Effectiveness of Reflection Content}
\label{exp:append_qa}
To study the effectiveness of the reflective content produced during MedReflect generation, we collected the reflection questions $R_Q$, reflection answers $R_A$, and reflection pairs $R_{QA}$ generated by the MedReflect-7B model, and provided them as additional context to the base Qwen2.5-7B-Instruct model. The results are shown in Figure~\ref{fig:appendqa}, and additional experimental details are provided in Appendix~\ref{appendix:appendqa}. As illustrated in Figure~\ref{fig:appendqa}, overall, augmenting the original Qwen2.5-7B-Instruct model with reflection information consistently obtains better performance than the base model alone, indicating that the reflections produced by MedReflect are beneficial. On highly specialized datasets such as GPQA, both reflection questions and reflection answers lead to substantial improvements. Moreover, using only reflection questions (without answers) still brings consistent gains, suggesting that MedReflect’s reflection questions help steer the direction of reasoning.
\begin{figure}[t]
    \centering
    \includegraphics[width=0.9\linewidth]{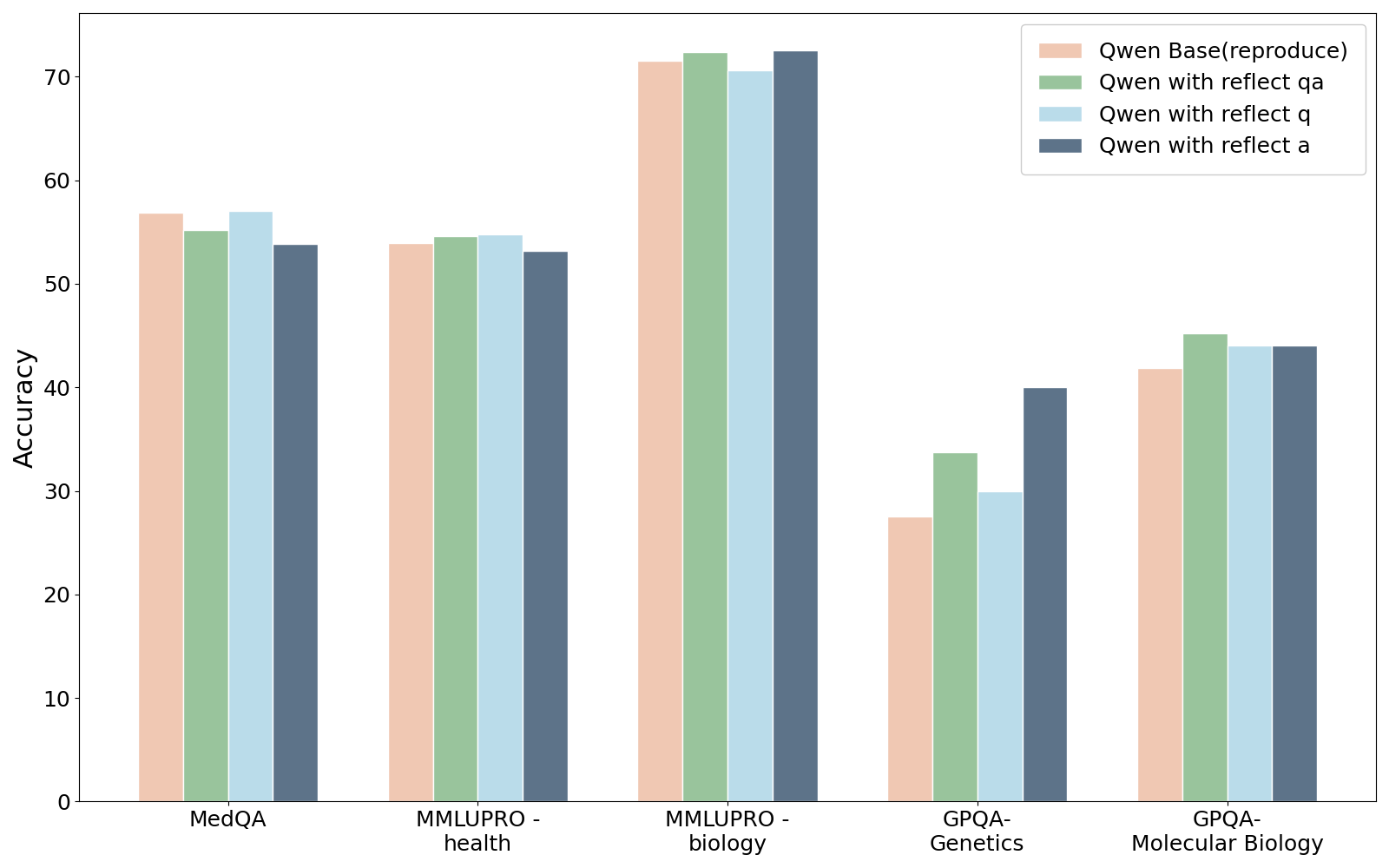}
    \caption{Results of the Attribution Analysis Experiment}
    \label{fig:appendqa}
\end{figure}

\section{Conclusion}
In this study, we propose MedReflect, a framework designed to enable LLMs to autonomously perform reflection and revision during medical tasks. By leveraging a lightweight LLM to construct a low-cost, diverse reflection training dataset, we trained models to acquire physician-like reflective thinking characterized. This training paradigm itself exhibits conspicuous cost-effectiveness. Notably, fine-grained reflection design proves crucial for fully activating the model’s reflective capabilities. Finally, we demonstrate that MedReflect significantly enhances model performance across multiple medical QA benchmarks, robustly validating the efficacy of the reflection mechanism in improving LLM accuracy in medical tasks. 
\newpage

\section*{Limitation}

While MedReflect significantly enhances diagnostic accuracy by mimicking the physician's reflective thinking process, this performance gain comes with an inherent trade-off in inference efficiency. As described in our method, the framework explicitly inserts a reflection chain before generating the final modified decision. Consequently, this mechanism inevitably increases the total number of generated tokens per query compared to direct-answer models. The increased sequence length results in higher computational costs and longer inference latency, which may pose challenges for deployment in real-time or resource-constrained medical consultation scenarios. Future work could explore methods to trigger reflection adaptively only when the model detects high uncertainty, thereby balancing accuracy and efficiency.

\section*{Ethics Statement}
MedReflect is built upon the Qwen2.5-Instruct architecture and, like all Large Language Models (LLMs), inherits intrinsic limitations such as the susceptibility to hallucinations and the potential generation of counterfactual medical advice. In high-stakes medical decision-making scenarios, any inaccuracies or misinterpretations of clinical narratives could lead to adverse outcomes.

Therefore, MedReflect is intended solely for research purposes to explore the potential of autonomous reflective thinking in medical AI. It is not designed for direct clinical deployment without expert human oversight. Researchers and developers utilizing this framework must be cognizant of these risks and implement robust safeguards, such as rigorous secondary validation loops. We further declare that the training data derived from public datasets (ChatDoctor and MedMCQA) was processed strictly for academic research, adhering to the data filtering protocols described in this work to ensure quality and privacy compliance.

\bibliography{custom}

@article{chen2024huatuogpt,
  title={Huatuogpt-o1, towards medical complex reasoning with llms},
  author={Chen, Junying and Cai, Zhenyang and Ji, Ke and Wang, Xidong and Liu, Wanlong and Wang, Rongsheng and Hou, Jianye and Wang, Benyou},
  journal={arXiv preprint arXiv:2412.18925
        
        
        
        
        
        
        
        
        
        
        
        
        
        
        
        },
  year={2024}
}

@article{chen2023huatuogpt,
  title={Huatuogpt-ii, one-stage training for medical adaption of llms},
  author={Chen, Junying and Wang, Xidong and Ji, Ke and Gao, Anningzhe and Jiang, Feng and Chen, Shunian and Zhang, Hongbo and Song, Dingjie and Xie, Wenya and Kong, Chuyi and others},
  journal={arXiv preprint arXiv:2311.09774
        
        
        
        
        
        
        
        
        
        
        
        
        
        
        
        },
  year={2023}
}

@article{team2025gemma,
  title={Gemma 3 technical report},
  author={Team, Gemma and Kamath, Aishwarya and Ferret, Johan and Pathak, Shreya and Vieillard, Nino and Merhej, Ramona and Perrin, Sarah and Matejovicova, Tatiana and Ram{\'e}, Alexandre and Rivi{\`e}re, Morgane and others},
  journal={arXiv preprint arXiv:2503.19786
        
        
        
        
        
        
        
        
        
        
        
        
        
        
        
        },
  year={2025}
}

@article{zhang2024ultramedical,
  title={Ultramedical: Building specialized generalists in biomedicine},
  author={Zhang, Kaiyan and Zeng, Sihang and Hua, Ermo and Ding, Ning and Chen, Zhang-Ren and Ma, Zhiyuan and Li, Haoxin and Cui, Ganqu and Qi, Biqing and Zhu, Xuekai and others},
  journal={Advances in Neural Information Processing Systems},
  volume={37},
  pages={26045--26081},
  year={2024}
}

@article{labrak2024biomistral,
  title={Biomistral: A collection of open-source pretrained large language models for medical domains},
  author={Labrak, Yanis and Bazoge, Adrien and Morin, Emmanuel and Gourraud, Pierre-Antoine and Rouvier, Mickael and Dufour, Richard},
  journal={arXiv preprint arXiv:2402.10373
        
        
        
        
        
        
        
        
        
        
        
        
        
        
        
        
        
        
        
        },
  year={2024}
}

@article{ye2024physics,
  title={Physics of language models: Part 2.2, how to learn from mistakes on grade-school math problems},
  author={Ye, Tian and Xu, Zicheng and Li, Yuanzhi and Allen-Zhu, Zeyuan},
  journal={arXiv preprint arXiv:2408.16293
        
        
        
        
        
        
        
        
        
        },
  year={2024}
}

@article{gandhi2024stream,
  title={Stream of Search (SoS): Learning to Search in Language},
  author={Gandhi, Kanishk and Lee, Denise and Grand, Gabriel and Liu, Muxin and Cheng, Winson and Sharma, Archit and Goodman, Noah D},
  journal={CoRR},
  year={2024}
}

@article{qin2025backtrack,
  title={To backtrack or not to backtrack: When sequential search limits model reasoning},
  author={Qin, Tian and Alvarez-Melis, David and Jelassi, Samy and Malach, Eran},
  journal={arXiv preprint arXiv:2504.07052
        
        
        
        
        
        },
  year={2025}
}

@article{xi2024enhancing,
  title={Enhancing LLM Reasoning via Critique Models with Test-Time and Training-Time Supervision},
  author={Xi, Zhiheng and Yang, Dingwen and Huang, Jixuan and Tang, Jiafu and Li, Guanyu and Ding, Yiwen and He, Wei and Hong, Boyang and Dou, Shihan and Zhan, Wenyu and others},
  journal={CoRR},
  year={2024}
}

@article{qu2024recursive,
  title={Recursive introspection: Teaching language model agents how to self-improve},
  author={Qu, Yuxiao and Zhang, Tianjun and Garg, Naman and Kumar, Aviral},
  journal={Advances in Neural Information Processing Systems},
  volume={37},
  pages={55249--55285},
  year={2024}
}

@misc{wu2024medicalgraphragsafe,
      title={Medical Graph RAG: Towards Safe Medical Large Language Model via Graph Retrieval-Augmented Generation}, 
      author={Junde Wu and Jiayuan Zhu and Yunli Qi and Jingkun Chen and Min Xu and Filippo Menolascina and Vicente Grau},
      year={2024},
      eprint={2408.04187},
      archivePrefix={arXiv},
      primaryClass={cs.CV},
      url={https://arxiv.org/abs/2408.04187}, 
}

@misc{lu2025doctorragmedicalragfusing,
      title={DoctorRAG: Medical RAG Fusing Knowledge with Patient Analogy through Textual Gradients}, 
      author={Yuxing Lu and Gecheng Fu and Wei Wu and Xukai Zhao and Sin Yee Goi and Jinzhuo Wang},
      year={2025},
      eprint={2505.19538},
      archivePrefix={arXiv},
      primaryClass={cs.CL},
      url={https://arxiv.org/abs/2505.19538}, 
}

@article{jin2021disease,
  title={What disease does this patient have? a large-scale open domain question answering dataset from medical exams},
  author={Jin, Di and Pan, Eileen and Oufattole, Nassim and Weng, Wei-Hung and Fang, Hanyi and Szolovits, Peter},
  journal={Applied Sciences},
  volume={11},
  number={14},
  pages={6421},
  year={2021},
  publisher={MDPI}
}

@inproceedings{pal2022medmcqa,
  title={Medmcqa: A large-scale multi-subject multi-choice dataset for medical domain question answering},
  author={Pal, Ankit and Umapathi, Logesh Kumar and Sankarasubbu, Malaikannan},
  booktitle={Conference on health, inference, and learning},
  pages={248--260},
  year={2022},
  organization={PMLR}
}

@article{jin2019pubmedqa,
  title={Pubmedqa: A dataset for biomedical research question answering},
  author={Jin, Qiao and Dhingra, Bhuwan and Liu, Zhengping and Cohen, William W and Lu, Xinghua},
  journal={arXiv preprint arXiv:1909.06146},
  year={2019}
}

@article{wang2024mmlu,
  title={Mmlu-pro: A more robust and challenging multi-task language understanding benchmark},
  author={Wang, Yubo and Ma, Xueguang and Zhang, Ge and Ni, Yuansheng and Chandra, Abhranil and Guo, Shiguang and Ren, Weiming and Arulraj, Aaran and He, Xuan and Jiang, Ziyan and others},
  journal={Advances in Neural Information Processing Systems},
  volume={37},
  pages={95266--95290},
  year={2024}
}

@inproceedings{rein2024gpqa,
  title={Gpqa: A graduate-level google-proof q\&a benchmark},
  author={Rein, David and Hou, Betty Li and Stickland, Asa Cooper and Petty, Jackson and Pang, Richard Yuanzhe and Dirani, Julien and Michael, Julian and Bowman, Samuel R},
  booktitle={First Conference on Language Modeling},
  year={2024}
}

@article{hu2022lora,
  title={Lora: Low-rank adaptation of large language models.},
  author={Shen, Yelong and Wallis, Phillip and Allen-Zhu, Zeyuan and Li, Yuanzhi and Wang, Shean and others},
    year={2022}
}

@article{LIN2025100868,
title = {Roles and Potential of Large Language Models in Healthcare: A Comprehensive Review},
journal = {Biomedical Journal},
pages = {100868},
year = {2025},
issn = {2319-4170},
doi = {https://doi.org/10.1016/j.bj.2025.100868},
url = {https://www.sciencedirect.com/science/article/pii/S2319417025000423},
author = {Chihung Lin and Chang-Fu Kuo}
}

@article{xu2025knowledge,
  title={Knowledge fusion in deep learning-based medical vision-language models: A review},
  author={Xu, Dexuan and Chen, Yanyuan and Chai, Zhongyan and Xiao, Yifan and Yan, Yandong and Ding, Weiping and Wang, Hanpin and Jin, Zhi and Jiao, Wenpin and Yue, Weihua and others},
  journal={Information Fusion},
  pages={103455},
  year={2025},
  publisher={Elsevier}
}

@inproceedings{dou-etal-2024-detection,
    title = "Detection, Diagnosis, and Explanation: A Benchmark for {C}hinese Medial Hallucination Evaluation",
    author = "Dou, Chengfeng  and
      Zhang, Ying  and
      Chen, Yanyuan  and
      Jin, Zhi  and
      Jiao, Wenpin  and
      Zhao, Haiyan  and
      Huang, Yu",
    editor = "Calzolari, Nicoletta  and
      Kan, Min-Yen  and
      Hoste, Veronique  and
      Lenci, Alessandro  and
      Sakti, Sakriani  and
      Xue, Nianwen",
    booktitle = "Proceedings of the 2024 Joint International Conference on Computational Linguistics, Language Resources and Evaluation (LREC-COLING 2024)",
    month = may,
    year = "2024",
    address = "Torino, Italia",
    publisher = "ELRA and ICCL",
    url = "https://aclanthology.org/2024.lrec-main.428/",
    pages = "4784--4794"
}

@inproceedings{kwon2024large,
  title={Large language models are clinical reasoners: Reasoning-aware diagnosis framework with prompt-generated rationales},
  author={Kwon, Taeyoon and Ong, Kai Tzu-iunn and Kang, Dongjin and Moon, Seungjun and Lee, Jeong Ryong and Hwang, Dosik and Sohn, Beomseok and Sim, Yongsik and Lee, Dongha and Yeo, Jinyoung},
  booktitle={Proceedings of the AAAI conference on artificial intelligence},
  volume={38},
  number={16},
  pages={18417--18425},
  year={2024}
}

@article{vladika2024medreqal,
  title={MedREQAL: Examining medical knowledge recall of large language models via question answering},
  author={Vladika, Juraj and Schneider, Phillip and Matthes, Florian},
  journal={arXiv preprint arXiv:2406.05845},
  year={2024}
}

@article{guo2025deepseek,
  title={Deepseek-r1: Incentivizing reasoning capability in llms via reinforcement learning},
  author={Guo, Daya and Yang, Dejian and Zhang, Haowei and Song, Junxiao and Zhang, Ruoyu and Xu, Runxin and Zhu, Qihao and Ma, Shirong and Wang, Peiyi and Bi, Xiao and others},
  journal={arXiv preprint arXiv:2501.12948},
  year={2025}
}

@article{team2024gemini,
  title={Gemini 1.5: Unlocking multimodal understanding across millions of tokens of context},
  author={Team, Gemini and Georgiev, Petko and Lei, Ving Ian and Burnell, Ryan and Bai, Libin and Gulati, Anmol and Tanzer, Garrett and Vincent, Damien and Pan, Zhufeng and Wang, Shibo and others},
  journal={arXiv preprint arXiv:2403.05530},
  year={2024}
}

@article{achiam2023gpt,
  title={Gpt-4 technical report},
  author={Achiam, Josh and Adler, Steven and Agarwal, Sandhini and Ahmad, Lama and Akkaya, Ilge and Aleman, Florencia Leoni and Almeida, Diogo and Altenschmidt, Janko and Altman, Sam and Anadkat, Shyamal and others},
  journal={arXiv preprint arXiv:2303.08774},
  year={2023}
}

@misc{li2023chatdoctormedicalchatmodel,
      title={ChatDoctor: A Medical Chat Model Fine-Tuned on a Large Language Model Meta-AI (LLaMA) Using Medical Domain Knowledge}, 
      author={Yunxiang Li and Zihan Li and Kai Zhang and Ruilong Dan and Steve Jiang and You Zhang},
      year={2023},
      eprint={2303.14070},
      archivePrefix={arXiv},
      primaryClass={cs.CL},
      url={https://arxiv.org/abs/2303.14070}, 
}

@article{rios2024evaluation,
  title={Evaluation of LLMs as a Diagnostic Aid for Complex Medical Cases},
  author={Ríos-Hoyo, A. and Shan, N.L. and Li, A.},
  journal={arXiv preprint arXiv:2410.xxxxx}, 
  year={2024}
}

@article{soffer2025pitfalls,
  title={Pitfalls of Large Language Models in Medical Ethics Reasoning},
  author={Soffer, S. and Sorin, V. and Nadkarni, G.N.},
  journal={arXiv preprint},
  year={2025}
}

@article{liu2025medical,
  title={Medical Large Language Model for Diagnostic Reasoning Across Specialties},
  author={Liu, X. and others},
  journal={arXiv preprint},
  year={2025}
}

@article{wu2025medreason,
  title={MedReason: Eliciting Factual Medical Reasoning Steps in LLMs via Knowledge Graphs},
  author={Wu, Juncheng and Deng, Wenlong and Li, Xingxuan and others},
  journal={arXiv preprint arXiv:2504.00993},
  year={2025}
}

@article{chen2025evaluating,
  title={Evaluating and Improving Syndrome Differentiation Thinking Ability in LLMs},
  author={Chen, C. and Wang, X. and Guan, M.},
  journal={arXiv preprint},
  year={2025}
}

@inproceedings{li2025reflectevo,
  title={Reflectevo: Improving meta introspection of small llms by learning self-reflection},
  author={Li, Jiaqi and Dong, Xinyi and Liu, Yang and Yang, Zhizhuo and Wang, Quansen and Wang, Xiaobo and Zhu, Song-Chun and Jia, Zixia and Zheng, Zilong},
  booktitle={Findings of the Association for Computational Linguistics: ACL 2025},
  pages={16948--16966},
  year={2025}
}

@article{zhou2025enhancing,
  title={Enhancing the Medical Context-Awareness Ability of LLMs via Multifaceted Self-Refinement Learning},
  author={Zhou, Yuxuan and Wang, Yubin and Wang, Bin and Ning, Chen and Liu, Xien and Wu, Ji and Hao, Jianye},
  journal={arXiv preprint arXiv:2511.10067},
  year={2025}
}

@article{li2025medfact,
  title={MedFact-R1: Towards Factual Medical Reasoning via Pseudo-Label Augmentation},
  author={Li, Gengliang and Chen, Rongyu and Li, Bin and Yang, Linlin and Ding, Guodong},
  journal={arXiv preprint arXiv:2509.15154},
  year={2025}
}

@article{jeong2024improving,
  title={Improving medical reasoning through retrieval and self-reflection with retrieval-augmented large language models},
  author={Jeong, Minbyul and Sohn, Jiwoong and Sung, Mujeen and Kang, Jaewoo},
  journal={Bioinformatics},
  volume={40},
  number={Supplement\_1},
  pages={i119--i129},
  year={2024},
  publisher={Oxford University Press}
}

@article{xu2024sayself,
  title={Sayself: Teaching llms to express confidence with self-reflective rationales},
  author={Xu, Tianyang and Wu, Shujin and Diao, Shizhe and Liu, Xiaoze and Wang, Xingyao and Chen, Yangyi and Gao, Jing},
  journal={arXiv preprint arXiv:2405.20974},
  year={2024}
}

@inproceedings{ji2023towards,
  title={Towards mitigating LLM hallucination via self reflection},
  author={Ji, Ziwei and Yu, Tiezheng and Xu, Yan and Lee, Nayeon and Ishii, Etsuko and Fung, Pascale},
  booktitle={Findings of the Association for Computational Linguistics: EMNLP 2023},
  pages={1827--1843},
  year={2023}
}

@article{kamoi2024can,
  title={When can llms actually correct their own mistakes? a critical survey of self-correction of llms},
  author={Kamoi, Ryo and Zhang, Yusen and Zhang, Nan and Han, Jiawei and Zhang, Rui},
  journal={Transactions of the Association for Computational Linguistics},
  volume={12},
  pages={1417--1440},
  year={2024},
  publisher={MIT Press 255 Main Street, 9th Floor, Cambridge, Massachusetts 02142, USA~…}
}

@article{zhu2025emergence,
  title={From Emergence to Control: Probing and Modulating Self-Reflection in Language Models},
  author={Zhu, Xudong and Jiang, Jiachen and Khalili, Mohammad Mahdi and Zhu, Zhihui},
  journal={arXiv preprint arXiv:2506.12217},
  year={2025}
}

@article{ye2025emergence,
  title={On the emergence of thinking in llms i: Searching for the right intuition},
  author={Ye, Guanghao and Pham, Khiem Duc and Zhang, Xinzhi and Gopi, Sivakanth and Peng, Baolin and Li, Beibin and Kulkarni, Janardhan and Inan, Huseyin A},
  journal={arXiv preprint arXiv:2502.06773},
  year={2025}
}
\newpage

\appendix

\section{Implementation Details for Data Generation}
\label{app:detail}
As outlined in the Methods section, we utilized the MedMCQA dataset to construct reflections at the sentence level and the ChatDoctor dataset to build reflections at the word level. The implementation details of this process are described below, with all corresponding prompts provided in Table \ref{tab:prompt_part1} and Table \ref{tab:prompt_part2}.
\paragraph{Implementation Detail for reflect pinpoint generation}

\paragraph{RG1: MedMCQA}
Given a multiple-choice question, we employ Prompt I to a LLM, eliciting both an initial response and its reasoning process. The model is required to strictly adhere to a predefined format specification for its final answer. Subsequently, we compare the model's generated final answer option against the standard answer: samples where the options differ are retained, while those that match are discarded. Then, using Prompt II, we identify where in the explanation the erroneous option manifests, isolating the corresponding segment of text containing the critical reasoning flaw. This segment serves as a pinpointed reflection target.
\paragraph{RG2: ChatDoctor}
Following text preprocessing, we employ Prompt III to guide the model in performing the entity extraction task. Subsequently, we randomly select 1 to 3 recognized entities within a sentence to mask, replacing each with entity type. We then utilize Prompt IV to instruct the model to fill these masked entities. Through multiple retries, we calculate a post-retry similarity score. Entities yielding a score below 0.8 are deemed incorrectly after retries and are retained in their masked state entity type. We repeat this step five times to obtain diverse data.

\paragraph{Implementation Detail for Reflection Generation}
The generation of reflection questions and answers then proceeds as follows:
Generating Reflection Questions: The erroneous masked entity  along with the original question are input to the model using Prompt V to generate guiding reflection questions. These questions specifically focus on the relationship between the erroneous entity and the correct answer.

Generating Reflective Answers: Prompt VI is used to direct the model to generate reflective answers based on its own knowledge and capabilities, addressing the questions generated in the previous step.

Generating Corrected Candidate Answers: Building on the reflection content from steps 1 and 2, Prompt VII guides the model to produce multiple candidate answers incorporating corrected entities.

Each generation step incorporates a retry mechanism to ensure valid output. Finally, the corrected candidate answers undergo strict validation requiring inclusion of all ground-truth correct entities to be accepted.

\paragraph{Data Filtering}
To guarantee the quality of the reflection data, we first preprocessed the raw datasets to remove samples with short reasoning paths or content irrelevant to medical question-answering. Following this, we conducted a secondary quality assessment to mitigate potential confirmation bias in the model-generated reflections.

This process entailed feeding the generated reflection questions ($R_q$) and answers ($R_a$), together with the model’s initial erroneous trajectory, back into the model to generate a revised answer. Crucially, the validity of a reflection was determined by verifying the revised answer against the objective ground truth provided by the datasets, ensuring that the correction was not merely endorsed by the model itself but aligned with the gold standard:

(1) For RG1, we utilized the ground-truth option of the multiple-choice question as the standard. A reflection instance was considered valid only if it guided the model to select the correct option matching the dataset label.

(2) For RG2, the objective was to restore masked medical entities. A reflection was deemed valid if the model, guided by the reflection, correctly predicted the original entity present in the raw medical text.

To further ensure the robustness of the reflection data, we repeated this evaluation process 10 times for each instance. We retained a reflection instance only if the model successfully reached the correct ground truth in at least 8 out of these 10 trials. Instances failing to meet this threshold were filtered out to exclude unstable or stochastic reasoning paths. Ultimately, this filtering process yielded a dataset comprising 36,413 medical consultation records and 21,107 multiple-choice questions.

\begin{table*}[htbp!]
\centering
\small
\renewcommand{\arraystretch}{1.3}
\begin{tabular}{l c c c c c c}
\toprule
\multirow{2}{*}{\textbf{Metric}} &
\multirow{2}{*}{\textbf{MedQA}} &
\multirow{2}{*}{\textbf{PubMedQA}} &
\multicolumn{2}{c}{\textbf{MMLU-Pro}} &
\multicolumn{2}{c}{\textbf{GPQA}}  \\
\cmidrule(lr){4-5} \cmidrule(lr){6-7}
& & & \textbf{Health} & \textbf{Biology} & \textbf{Genetics} & \textbf{\makecell{Molecular\\Biology}} \\ 
\midrule
\textbf{Avg. Generated Tokens} (All) & 711.18 & 321.31 & 641.96 & 721.19 & 840.35 & 730.50 \\
\textbf{Avg. Generated Tokens} (Correct)  & 711.39 & 320.14 & 638.30 & 674.78 & 904.73 & 697.90 \\
\textbf{Avg. Generated Tokens} (Incorrect) & 710.90 & 322.20 & 645.94 & 810.58 & 761.67 & 761.58 \\
\bottomrule
\end{tabular}
\caption{Statistics of average response length (number of tokens) across different medical benchmarks.}
\label{tab:response_length}
\end{table*}

\section{Dataset Analysis}
\label{app:data}
\begin{table}[h!]

  \centering
  \small
  \begin{tabular}{lcccc}
    \hline
    dataset    & Num    & Total\_reflections  & Length \\
    \hline
    MedMCQA       & 21107  & 21107               & 394.56        \\
    ChatDoctor & 36413  & 55460              & 305.05        \\
    \hline
  \end{tabular}
  \caption{Data Analysis}
  \label{tab:dataset_stats}
\end{table}
The key datasets underpinning this study are MedMCQA and ChatDoctor. As shown in the table \ref{tab:dataset_stats}, each sample in the MedMCQA dataset (21,107 samples) contains only a single reflection instance (Reflect avg = 1.0), with reflections primarily occurring at the sentence level. In contrast, the ChatDoctor dataset is significantly larger (36,413 samples) and features richer, more granular reflection content. It encompasses a total of 55,460 reflection instances, averaging 1.52 reflections per sample. This indicates a substantial number of dialogues within ChatDoctor involve diverse reflective processes. Furthermore, ChatDoctor focuses specifically on reflecting medical entities, enabling precise identification of cognitive biases at the entity level and capable of generating 1 to 3 distinct reflection points per response. This structural disparity between the datasets provides the essential data foundation for training models to understand and generate reflections at varying levels of granularity.
\begin{table*}[h] 
    \centering
    \renewcommand{\arraystretch}{1.2} 
    \begin{tabular}{|p{0.15\textwidth}|p{0.8\textwidth}|} 
        \hline
        \textbf{ID} & \textbf{Detail} \\
        \hline
        Prompt I & You are a medical expert skilled in multiple choice question answering with thorough reasoning.
        $<$Query$>$:[Question]$<$/Query$>$\newline
        If you need to specify an option, always use option A/B/C/D.
        Now, conclude your answer [Therefore, my answer is: option A/B/C/D.] after thorough reasoning: \\
        \hline
        Prompt II & You are a text processor skill in spliting text according to the option.
        Split the text into prefix and suffix, the prefix is the content the first time the option is explained in the text. Do not alter the original text.
        $<$Text$>$[Text to be segmented]$<$/Text$>$\newline
        $<$Option$>$[Incorrect option letter]$<$/Option$>$\newline
        $<$Prefix$>$ \\
        \hline
        Prompt III & You are a professional doctor. I will provide you with a question and part of the answer. Please read the answer and extract the relevant entities that appear in the answer. Extract professional core terms with high clinical medical characteristics from the perspective of professional doctors, such as etiology, disease, diagnosis, medical examination and testing, medicine, medical therapy and other medical entities. Find all the medical entities that you think suitable in the answer. If you think there are no suitable entities, just ok to return null. Please provide it to me in JSON format, such as [entity type:xxx,entity name:xxx]. No explanation is needed, just give me JSON in English.\\
        \hline
        Prompt IV & Here is a medical query from your patient:\newline$<$Query$>$:[question]$<$/Query$>$\newline Here is the answer template for the medical query:\newline$<$Answer template$>$:[temp text]$<$/Answer template$>$\newline Please complete the mask section based on the template I provided. The case format for the mask section is as follows:  $<$ mask, type: TYPE$>$, The TYPE is a hint I gave you, implying the entity type in the mask section. Each part of the mask can only be filled with one answer, which can be a word or phrase. You need to find all the mask entities in this sentence and give me the answer you generated, with $<$mask, type: TYPE, answer: YOURANSWER$>$ is returned without explanation. The TYPE must be the same as the TYPE I gave you. Just give me the return format. Your must give me a answer. And it cannot be the same as the TYPE. Now you can answer here:\\
        \hline
        Prompt V & Here is a medical query from your patient:\newline$<$Query$>$:[query]$<$/Query$>$\newline
        Here is your response for the medical query:\newline$<$Response$>$:[sentences]$<$/Response$>$\newline
        ($<$correct$>$ [your correct preceding content] $<$/correct$>$,$<$incorrect$>$[your wrong answer] $<$/incorrect$>$)\newline
        The initial answer [retry answer] in your response is incorrect, so you need to ask a reflective question based on your correct preceding content (if any) and your wrong answer. \newline Now please provide a brief question(Strictly follow this format:$<$Reflective Question$>$your response$<$/Reflective Question$>$):\\
        \hline
        
    \end{tabular}
    \caption{Prompts for Data Generation (Part 1)}
    \label{tab:prompt_part1}
\end{table*}

\begin{table*}[h]
    \centering
    \renewcommand{\arraystretch}{1.2}
    \begin{tabular}{|p{0.15\textwidth}|p{0.8\textwidth}|}
        \hline
        \textbf{ID} & \textbf{Detail} \\
        
        \hline
        Prompt VI & Here is a medical query from your patient:\newline$<$Query$>$:[query]$<$/Query$>$\newline
        Here is your response for the medical query:\newline$<$Response$>$:[sentences]$<$/Response$>$\newline
        Here is your own reflective question for your response:\newline$<$Reflective Question$>$:[Reflective Question]$<$/Reflective Question$>$\newline
        ($<$correct$>$ [your correct preceding content] $<$/correct$>$,$<$incorrect$>$[your wrong answer] $<$/incorrect$>$)\newline
        Now, please provide a concise answer for the reflective question(Strictly follow this format:$<$Reflective Answer$>$your response$<$/Reflective Answer$>$):\\
        \hline
        Prompt VII & Here is a medical query from your patient:\newline$<$Query$>$:[query]$<$/Query$>$\newline
        Here is your own reflection on your initial answer:\newline$<$Self-Reflection$>$:[reflect question][reflect answer]$<$/Self-Reflection$>$\newline
        Here is the response you need to complete(Complete each blank):\newline$<$Response$>$:[sentences]$<$/Response$>$\newline
        Now, according on this reflection, your completed answer is(Strictly follow this format:$<$Answer$>$:your refine entity,eg:[weight reduction,cancer]$<$/Answer$>$):\\
        \hline
    \end{tabular}
    \caption{Prompts for Data Generation (Part 2 - Continued)}
    \label{tab:prompt_part2}
\end{table*}

\section{More Implementation Details}
\label{app:appendix_implementation}

To ensure the reproducibility of MedReflect and address the specifics of our training and inference mechanisms, we provide detailed configurations of training.

\subsection{Training Details.}

\paragraph{Training Data Statistics} We tailored the scale of the fine-tuning dataset to the capacity of the target models to ensure optimal adaptation. Specifically, we utilized 2k training examples for the Qwen2.5-7B-Instruct model. For the larger Qwen2.5-32B-Instruct model, we scaled up the training set to 30k examples to prevent under-fitting and ensure robust performance.

\paragraph{Configuration of Supervised Fine-Tuning (SFT).}
We fine-tuned the base models (Qwen2.5-7B/32B-Instruct) using the generated MedReflect dataset. Table \ref{tab:hyperparameters} details the specific Configurations used in our experiments. We utilized the AdamW optimizer with a cosine learning rate scheduler. The training was performed on 4 $\times$ NVIDIA A800 (80GB) GPUs.

\begin{table}[H] 
    \centering

    \setlength{\tabcolsep}{3pt} 

    \begin{tabular}{lc}
        \toprule
        \textbf{Configuration} & \textbf{Value} \\
        \midrule
        Global Batch Size & 4 \\
        Learning Rate & $1e-4$ \\
        Epochs & 3 \\
        Max Sequence Length & 4096 \\
        Optimizer & AdamW \\
        Weight Decay & 0.01 \\
        Warmup Ratio & 0.03 \\
        LR Scheduler & Cosine \\
        Gradient Clipping & 1.0 \\
        \midrule
        \multicolumn{2}{l}{\textit{\textbf{LoRA Configuration}}} \\
        LoRA Rank ($r$) & 16 \\
        LoRA Alpha ($\alpha$) & 8 \\
        LoRA Dropout & 0.05 \\
        Target Modules & \texttt{q}, \texttt{k}, \texttt{v}, \texttt{o}, \\ 
                       & \texttt{gate}, \texttt{up}, \texttt{down\_proj} \\
        \bottomrule
    \end{tabular}
    \caption{Detailed configuration for SFT}
    \label{tab:hyperparameters}
\end{table}


\paragraph{Prompt Template}
To ensure the model adheres to the desired instruction-following format and medical reasoning style, we employed a specific template during the supervised fine-tuning stage. The input sequence is formatted as follows:

\begin{figure}[H]
    \centering
    \fbox{
    \begin{minipage}{0.9\linewidth}
        \ttfamily
        \small
        <|im\_start|>system \\
        You are a medical expert skilled in answer questions while reflecting. Your reflecting goal is to generate a good reflection to assist you to improve your previous answer and continue to answer the question. \\
        <|im\_end|> \\
        <|im\_start|>user \\
        \{Input Question\} \\
        <|im\_end|> \\
        <|im\_start|>assistant \\
        \{Target Response\} \\
        <|im\_end|>
    \end{minipage}
    }
    \caption{The training prompt template used for MedReflect.}
    \label{fig:prompt_template}
\end{figure}

\subsection{Evaluation Details}

\paragraph{Inference Configuration}
During the inference stage, we utilized the vllm library. We employed \textbf{Nucleus Sampling} to ensure the diversity and robustness of the generated responses. The specific generation parameters are listed in Table~\ref{tab:inference_params}.
\begin{table}[H]
    \centering

    \setlength{\tabcolsep}{6pt} 
    \begin{tabular}{lc}
        \toprule
        \textbf{Parameter} & \textbf{Value} \\
        \midrule
        Temperature & 0.7 \\ 
        Top-p  & 0.9 \\ 
        Repetition Penalty & 1.05 \\ 
        Max New Tokens & 4096 \\ 
        \bottomrule
    \end{tabular}
    \caption{Inference Configurations}
    \label{tab:inference_params}
\end{table}

\subsection{Evaluation Protocol}
To mitigate the variance arising from generation randomness and ensure the reliability of our results, we adopted a multi-run evaluation strategy. For standard benchmarks including MedQA, MMLU, PubMedQA, and MedMCQA, we conducted 3 independent runs and reported the average accuracy. For the GPQA dataset, given its relatively small test set size which is susceptible to higher statistical fluctuation, we increased the number of independent runs to 5 and reported the average performance to ensure a more stable assessment.

\subsection{Answer Extraction and Verification Strategy}
To standardize the evaluation process and mitigate discrepancies arising from formatting inconsistencies in model outputs, we implemented a robust two-stage protocol for answer extraction across all datasets.

\paragraph{Stage 1: Deterministic Pattern Matching.}
Initially, we employed a strict rule-based extraction mechanism using regular expressions. We defined a set of heuristic patterns to capture conclusive answers, such as matches for \texttt{"The answer is option [X]"} or \texttt{"My answer is [X]"}. If a distinct option key (A, B, C, or D) was successfully extracted, it was directly compared with the ground truth label.

\paragraph{Stage 2: LLM-Assisted Semantic Parsing.}
To address instances where the model's response structure was irregular or failed the regex matching (i.e., the answer was implicitly embedded in the reasoning chain without a standard prefix), we utilized a superior-capability LLM (GPT-4) as a semantic parser. This acts as a fallback mechanism to interpret the semantic intent of the model's output. 

Specifically, we utilized the \texttt{gpt-4-0613} version via API with a temperature setting of 0 to ensure deterministic and reproducible outputs. The model was tasked to act as an objective evaluator. We constructed a specific prompt template (detailed in Table~\ref{tab:eval_prompt}) that includes the original question, the candidate options, and the model's raw response. GPT-4 was instructed to identify the final chosen option and output a single character. This hybrid approach ensures that valid responses are not penalized due to minor formatting deviations, thereby providing a more accurate assessment of the model's reasoning capabilities.

\subsection{Evaluation setup of Experiment in Section~\ref{exp:append_qa}}
\label{appendix:appendqa}
In the experiments presented in Section~\ref{exp:append_qa}, the reflections (comprising reflection questions and answers) within the MedReflect-7B responses were extracted. This reflection information was subsequently utilized as context and concatenated with the original query to directly input into the original Qwen2.5-7B-Instruct. The prompt template used for this inference is shown in Table~\ref{tab:prompt_appendqa}.

\begin{table*}[h] 
    \centering
    \renewcommand{\arraystretch}{1.2} 
    \begin{tabular}{|p{0.15\textwidth}|p{0.8\textwidth}|} 
        \hline
        Prompt  & You are an experienced doctor. Please answer the following medical questions. Conclude your response with `Therefore, my answer is **your option letter**'. For example, if the your answer is option A, you should conclude your response with `Therefore, my answer is **A**.'.\newline
        $<$Query$>$:[Question]$<$/Query$>$\newline
        $<$Internal Thinking$>$[reflection\_info]$</$Internal Thinking$>$\newline [Warning: The $<$Internal Thinking$>$ represents your internal thoughts about the $<$Question$>$, it's not always correct, it's only for your reference.] \\
        \hline
        
    \end{tabular}
\caption{Prompts for the experiments in Section~\ref{exp:append_qa}. Depending on the experimental setting, \texttt{reflection\_info} may include both the reflection question and answer, only the reflection question, or only the reflection answer.}

    \label{tab:prompt_appendqa}
\end{table*}


\begin{table}[H]
    \centering
    \small 
    \renewcommand{\arraystretch}{1.4} 
    \begin{tabular}{p{0.2\linewidth} p{0.7\linewidth}}
        \toprule
        \textbf{Component} & \textbf{Content Details} \\
        \midrule
        \textbf{System Role} & 
        You are an excellent medical exam assistant, skilled at extracting correct information from complex responses. \\
        \midrule
        \textbf{Instruction} & 
        I will provide you with a question and the response from a large model. Please help determine which answer the large model believes is correct. If you think it does not answer the multiple-choice question, return None. No explanation is needed, just return the option itself.
                    question: {question}
                    answer: {ans}  \\
        \midrule
        \textbf{Input Format} & 
        \textbf{Question:} [Insert Question Text] \newline
        \textbf{Options:} \newline
        (A) [Option A] \quad (B) [Option B] \newline
        (C) [Option C] \quad (D) [Option D]  \\
        \bottomrule
    \end{tabular}
    \caption{The prompt template employed for the GPT-4 based answer extraction stage. The model acts as a semantic parser to extract the final answer key from unstructured reasoning chains.}
    \label{tab:eval_prompt}
\end{table}

\section{More Qualitative Analysis}
\subsection{Analysis of the Efficiency Boundary for Reflection Data}
\begin{table}[H]
    \centering
    \small

    \begin{tabular}{lcc}
        \toprule
        \textbf{Dataset} & \textbf{Reflect\_2k} & \textbf{Reflect\_30k} \\
        \midrule
        MedQA & 75.5 & 74.3\\
        PubMedQA & 75.3 & 74.0 \\
        MMLU-health & 62.8 & 61.0 \\
        MMLU-biology & 75.8 & 72.9 \\
        GPQA-Genetics & 65.0 & 63.0 \\
        GPQA-Molecular Biology & 60.3 & 59.5 \\
        \bottomrule
    \end{tabular}
    \caption{Results of the Data Efficiency Experiment of MedReflect-7B}
    \label{tab:model_performance}
\end{table}
We investigated the impact of training data magnitude on model performance by conducting a comparative analysis between a compact dataset of 2,000 samples and a significantly larger dataset of 30,000 samples. As indicated by the results shown in Table \ref{tab:model_performance}, the model trained on only 2,000 samples achieves a performance level comparable to that of the model trained on 30,000 samples. Across various metrics, the smaller dataset yields results commensurate with those of the larger dataset. This observation underscores the exceptional data efficiency of the proposed reflection mechanism, suggesting that the model can effectively internalize the core patterns of reflective reasoning with a limited volume of high-quality instruction data.

\subsection{Response Length.}
We further report the response length of MedReflect, measured by the number of generated tokens (excluding the prompt). Table~\ref{tab:response_length} summarizes the average length over all samples, as well as the average length of correct and incorrect responses. On MedQA and PubMedQA, correct and incorrect responses have almost the same average length. On MMLUPro-health, correct responses are slightly shorter on average. In contrast, on MMLUPro-biology and GPQA-Molecular Biology, incorrect responses are notably longer than correct ones. On the highly specialized GPQA-Genetics benchmark, correct responses are substantially longer than incorrect ones (904.7 vs.\ 761.7 tokens on average), suggesting a stronger length--correctness association on this dataset. Overall, response length appears dataset-dependent under MedReflect.

\section{Case Study}
\label{app:case}
\begin{figure*}[h]
    \centering
    \includegraphics[width=1\linewidth]{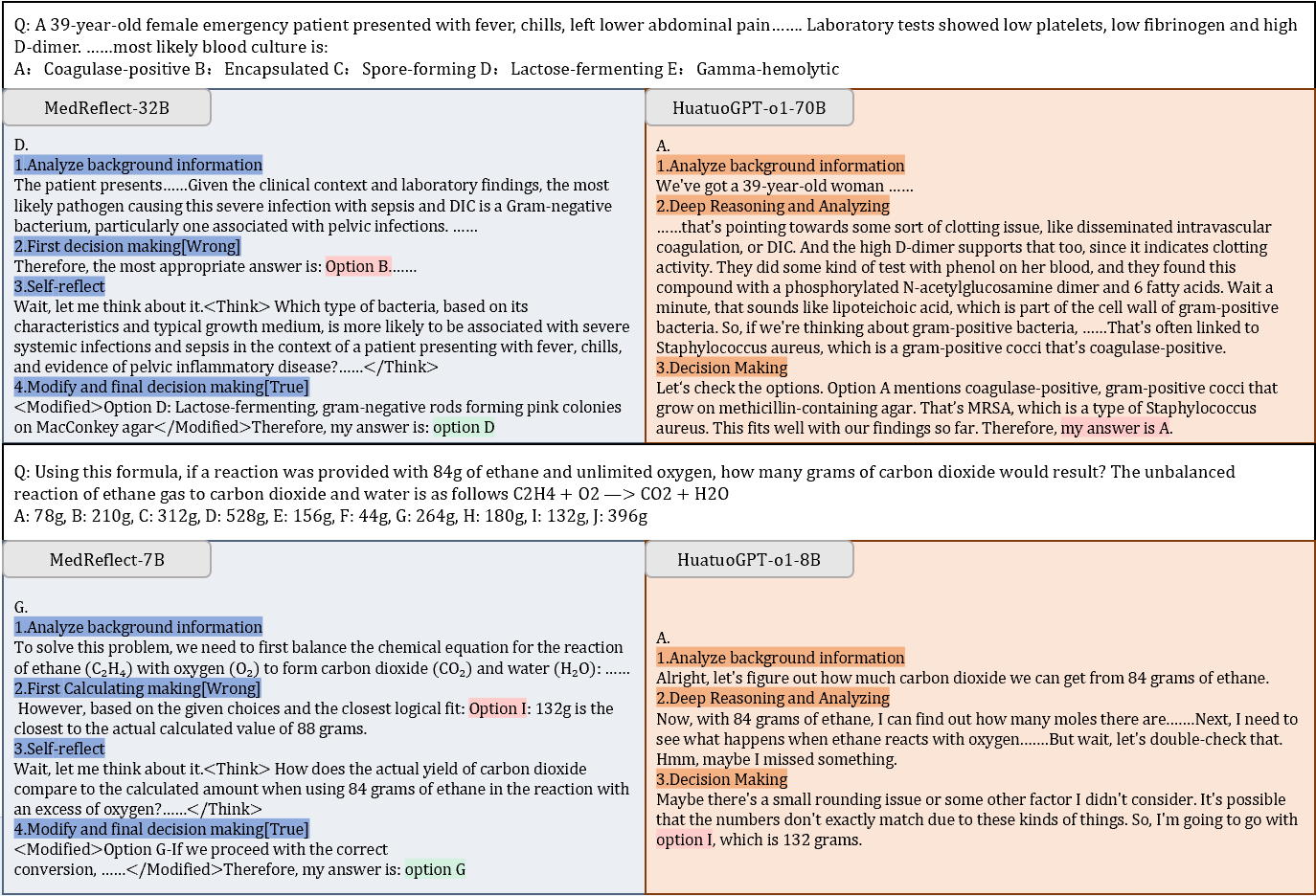}
    \caption{Cases of Multi-Choice Questions}
    \label{fig:multi-choice}
\end{figure*}

\paragraph{Cases for Multi-Choice Question}
The multi-choice cases are presented in Figure~\ref{fig:multi-choice}, where we compare the performance of MedReflect, Huatuogpt-o1, and Deepseek-R1.

\paragraph{I}

We conducted a specific analysis of the performance of MedReflect and the general reasoning large model (DeepSeek-R1), using septicemia pathogen identification as a representative clinical scenario. Experimental results reveal that both DeepSeek-R1 and MedReflect initially exhibit similar diagnostic errors when confronted with complex clinical evidence. However, a critical divergence emerges in their subsequent reasoning pathways:
\begin{itemize}
    \item DeepSeek-R1, constrained by its reasoning paradigm, terminates analysis upon reaching an incorrect conclusion. It demonstrates no self-correction, maintaining confidence in its initial output.
    \item MedReflect, leveraging its intrinsic reflective architecture, critically evaluates morphological inconsistencies – questioning the mismatch between observed coccobacilli features and Neisseria gonorrhoeae's typical diplococcal structure. It further assesses clinical epidemiology to validate diagnostic hypotheses.
\end{itemize}
This structured self-reflection mechanism enables MedReflect to iteratively retrieve evidence from the knowledge base and dynamically adjust diagnostic weights, thereby achieving autonomous error correction and precise diagnosis. In contrast, DeepSeek-R1 despite engaging in constant deep thinking, lack a mechanism for reflection and adjustment of initial responses.
\paragraph{II}

We conducted a specific analysis of the reasoning differences between MedReflect and Huatuogpt-o1, using mass conversion problems as a representative stoichiometric scenario.

\begin{itemize}
    \item MedReflect first incorrectly identified ethane.    Then, MedReflect exhibited dynamic error correction capability.   When significant deviations from given options were detected in its computed results, MedReflect implicitly switched to the correct molecular formula for recalculation, ultimately arriving at the correct answer (G).
    \item Huatuogpt-o1 manifests a fundamental flaw in its misconception of core principles,in which case it misinterprets the stoichiometric ratio.    This foundational error causes the entire computational framework to deviate from theoretical benchmarks.   Then, when the calculated result (122.4g) shows significant deviation from options, the model arbitrarily selects the closest match (I) without verifying the validity of stoichiometric assumptions, which indicates a lack of scientific verification or reflection.
\end{itemize}

\begin{figure*}[h]
    \centering
    \includegraphics[width=1\linewidth]{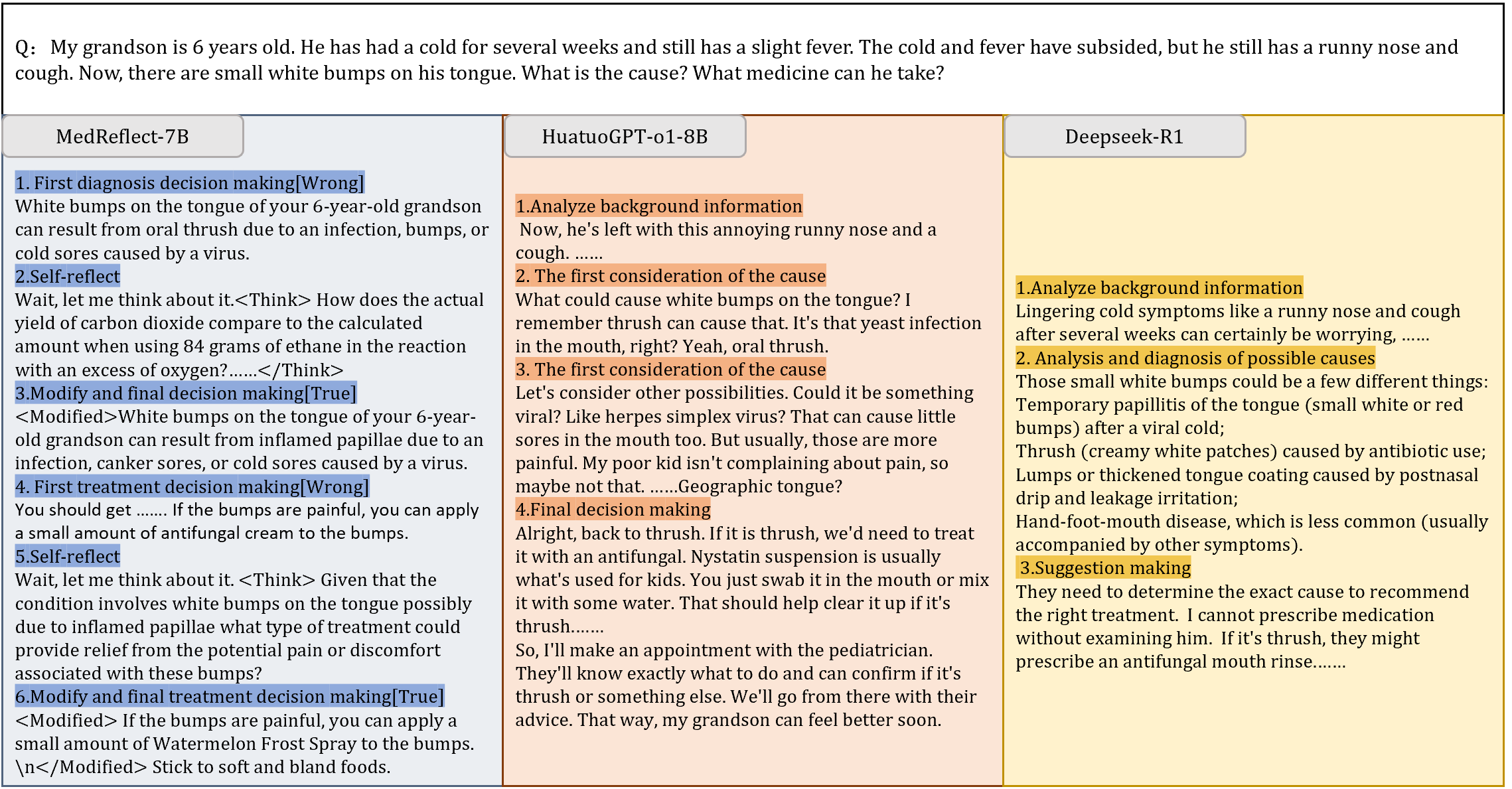}
    \caption{Case of Consulting Question}
    \label{fig:case-consult}
\end{figure*}

Overall, MedReflect tends to follow the physician-like reasoning process and adjust its reasoning process better. In contrast, HuatuoGPT-o1 failed to promptly correct its initial error, compounding the mistake until it ultimately settled for an answer it deemed a close approximation.
\paragraph{Cases for Consulting Question}
Using a diagnostic case of pediatric respiratory infection presenting with lingual symptoms, we evaluated the performance of three models: DeepSeek, MedReflect, and HuatuoGPT-o1. The performance is shown in Figure~\ref{fig:case-consult}.
\begin{itemize}
    \item MedReflect demonstrated more rigorous clinical reasoning through its dynamic reflection mechanism. After generating an initial response, it actively challenged its preliminary assumptions. This process involved critically analyzing relevant pathological mechanisms and prioritizing non-infectious etiologies. Its treatment recommendations appropriately advocated for conservative interventions such as avoiding irritants and dietary modifications.
    \item HuatuoGPT-o1 also engaged in deep reasoning and reflection during its analysis. However, its reflective process was comparatively superficial. Rather than conducting an in-depth exploration of disease etiology, it employed overly simplistic reasoning, such as excluding diagnoses based solely on "the child not complaining of pain".
    \item DeepSeek-R1 systematically listed potential diagnoses including lingual papillitis, oral thrush, and viral exanthems. However, its diagnostic suggestions exhibited a scattered approach, failing to effectively integrate core clinical clues. Of course, as a powerful general reasoning model, it may be difficult for it to directly offer suggestions, so it provides many options for patients to choose.
\end{itemize}
MedReflect's iterative reasoning model, grounded in clinical clue integration, differs significantly from DeepSeek's divergent deep reasoning. It more closely aligns with medical differential diagnostic logic. Simultaneously, its reflective process demonstrates greater rigor and professionalism compared to HuatuoGPT-o1's relatively simplistic and cursory approach.

\end{document}